\documentclass{article}
\usepackage[utf8]{inputenc}
\usepackage{graphicx}
\usepackage{multirow}
\usepackage{amsmath}
\usepackage{todonotes}
\usepackage{fancyhdr}
 
\usepackage[margin=1in]{geometry}

\title{Lifelong Learning Metrics}
\author{Alexander New$^*$,  Megan Baker, Eric Nguyen, Gautam Vallabha \\ Johns Hopkins University Applied Physics Laboratory\\ Laurel, MD 20723 \\ \texttt{alex.new@jhuapl.edu}, \texttt{megan.baker@jhuapl.edu},\\ \texttt{eric.nguyen@jhuapl.edu}, \texttt{gautam.vallabha@jhuapl.edu}\\
\small{$^*$corresponding author}}
\date{\today}
\begin{document}

\maketitle

\tableofcontents

\section{Introduction}

The DARPA Lifelong Learning Machines (L2M) program seeks to yield advances in artificial intelligence (AI) systems so that they are capable of learning (and improving) continuously, leveraging data on one task to improve performance on another, and doing so in a computationally sustainable way. Performers on this program developed systems capable of performing a diverse range of functions, including autonomous driving, real-time strategy, and drone simulation. These systems featured a diverse range of characteristics (e.g., task structure, lifetime duration), and an immediate challenge faced by the program's testing and evaluation team was measuring system performance across these different settings. This document, developed in close collaboration with DARPA and the program performers, outlines a formalism for constructing and characterizing the performance of agents performing lifelong learning scenarios.

In Section~\ref{sec:concepts}, we introduce the general form of a lifelong learning scenario. This requires specifying different types of experiences agents may be exposed to, and what metrics should be generated from these experiences. In Section~\ref{sec:conditions}, we briefly outline the criteria for an agent to demonstrate lifelong learning. In Section~\ref{sec:metrics}, we define a set of metrics that characterize to what extent an agent demonstrates lifelong learning in a given scenario. Discussions in sources such as~\cite{farquhar2019robust,Hsu2018categorization,vandeVen2019ThreeScenarios} are also useful for grounding the ideas behind lifelong learning.

Our framework and metrics are meant to be as agnostic as possible to the agent configuration (e.g., progressive network~\cite{Rusu2016ProgressiveNetwork} or elastic weight consolidation \cite{Kirkpatrick2016elastic}), domain (e.g., autonomous navigation, robotics, strategy, classification), and environment (e.g., StarCraft~\cite{StarCraft}, AirSim~\cite{Airsim2017fsr}, CARLA~\cite{Dosovitskiy17CARLA}, Habitat~\cite{habitat19iccv}, Arcade~\cite{bellemare13arcade}, SplitMNIST~\cite{Hsu2018categorization}, or Core50~\cite{Lomonco2017Core50}). It may also be used in combination with platforms for lifelong learning, such as Avalanche~\cite{Lomonaco2021Avalanche} or CORA~\cite{Powers2021CORA}.

Agents, domains, environments, and other terms are defined in more detail in Appendix~\ref{sec:glossary}. We recommend~\cite{Parisi2019LL} as an overview of recent approaches and advances in the general field of lifelong learning. Historically, there has been wide variation in how systems and metrics for lifelong learning are defined and assessed; different papers may focus on different metrics. Beyond those papers cited here, many others motivate their system designs with arguments to concepts like positive transfer. This document provides a consistently-defined suite of metrics applicable for general lifelong learning problems. In particular, although much early work in the field of lifelong learning focused on the problem of mitigating catastrophic forgetting~\cite{French1999CF},~\cite{McCloskey1989CF} -- an agent losing the previously-acquired ability to perform tasks as it encounters new tasks -- our metrics here endeavor to capture both catastrophic forgetting and other characterizations of lifelong learning, such as transfer and comparison to agents exposed to only a single task.

A Python library, \texttt{l2metrics}, containing implementations of these metrics is in development and will be publicly available shortly. This document will be updated when it is.

\subsection{How this document was developed}

Over the course of the DARPA L2M program, performers, the testing and evaluation team, and DARPA Systems Engineering and Technical Advisors (SETAs) formed several Working Groups that met regularly to discuss the concepts behind and metrics for characterizing Lifelong Learning. This document captures the consensuses arrived at after these discussions, and its content could only have been developed in the close collaboration this process entailed. In particular, the Definitions and Scenarios Working Group developed concepts like criteria of lifelong learning (Section~\ref{sec:conditions}), and tasks and environments (Section~\ref{sec:concepts}), the Metrics Working Group formulated the metrics (Section~\ref{sec:metrics}), and, during and after program evaluations, the performers gave feedback on the definitions and metrics based on their experiences using them in their systems.

\section{Concepts for Measuring Lifelong Learning}\label{sec:concepts}

\begin{figure}
    \centering
    \includegraphics[width=0.9\textwidth]{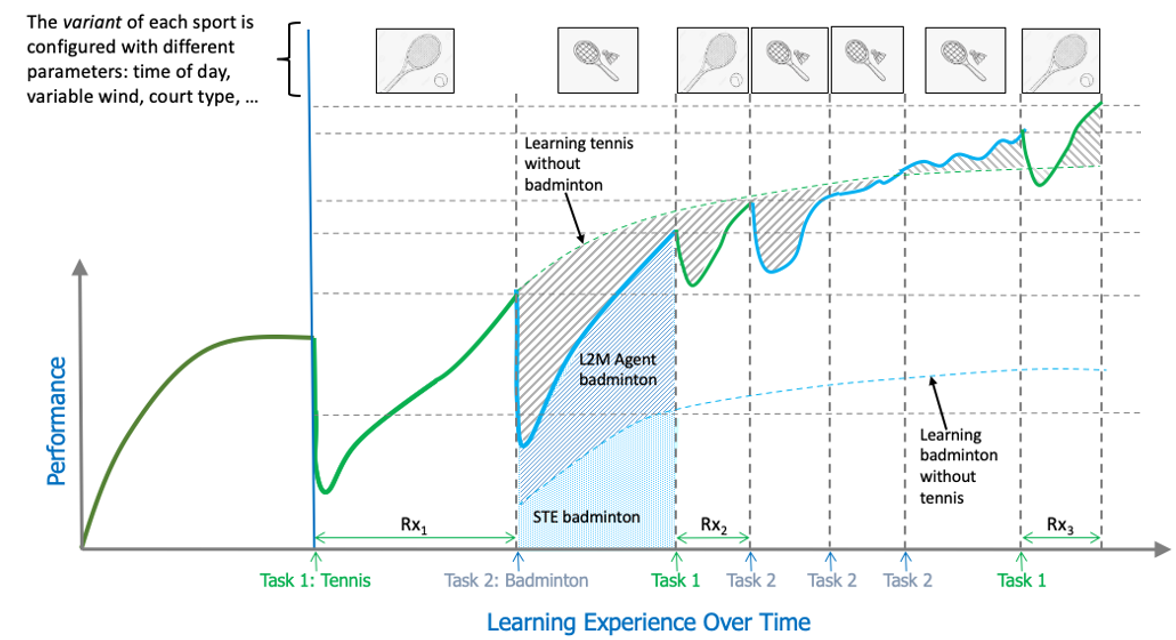}
    \caption{Notional overview of Lifelong Learning performance for an agent, showing performance on two tasks – tennis and badminton. Each time the L2M agent switches to a new task, its performance ($y$-axis) drops, but then it recovers. Its ability to leverage transfer between the tasks enables it to surpass the performance of an agent exposed only to one skill (dashed line).
    Here, $\text{Rx}_\text{i}$ denotes the length (in ``experiences'') of the $i$th occurrence of tennis in the agent's lifetime (its $i$th Learning Block), and STE refers to a Single Task Expert exposed only to badminton.}
    \label{fig:notional_overview}
\end{figure}

In this document, we define a \textbf{Lifelong Learning Agent} (L2 Agent) as an agent that continues learning once it is deployed. Here, \textbf{agent} refers to a learner that is interacting with an environment, where each action (output) of the agent shapes the agent’s future observations. The agent is able to adapt efficiently and flexibly to new tasks and new variations of known tasks; becomes a better learner over time (learn new tasks better and faster, perform better on old tasks); and is able to do these in a scalable and sustained manner. Feedback to the agent in response to its actions is given via occasional scalar “rewards”; however, the concepts discussed in this document are readily applicable to classification learners as well.

Fig.~\ref{fig:notional_overview} notionally describes the L2 concept, where an agent receives some initial training and then learns a sequence of tasks, gradually improving its performance over time. In principle, a lifelong learner should be able to learn a pre-specified task sequence (fixed curriculum), be able to select the tasks to learn and their order (based on criteria like novelty), handle overlapping or concurrent tasks, and so on. For current purposes, we focus on assessing the ability of a L2 agent to learn a pre-specified task sequence, with a mixture of old and new tasks. Although it is beyond the scope of this document, many of the concepts and measures introduced here may be extended to situations where multiple tasks are running simultaneously.

\subsection{Conditions of Lifelong Learning}\label{sec:conditions}

The following conditions were identified as being required for an agent to demonstrate lifelong learning. Each of the metrics in Section~\ref{sec:metrics} was developed to characterize an agent's ability to meet one of these criteria.

\begin{enumerate}
    \item \textbf{Continual Learning:} The L2M learns a nonstationary stream of tasks (both novel and recurring) without distinct training and testing phases, and continually consolidates new information to improve performance while coping with irrelevance and noise.
    
    \item \textbf{Transfer and Adaptation:} As learning progresses, the L2M performs better on average on the next task it experiences, for both novel and known tasks (forward and reverse transfer), and maintaining performance during rapid changes in the ongoing task (adaptation).
    
    \item \textbf{Sustainability:} The L2M continues learning for an arbitrarily long lifetime using limited resources (e.g., memory, time) in a scalable way.
\end{enumerate}

\subsection{Components of lifelong learning scenarios}

An L2 agent learns in a specific \textbf{environment}, which defines the set of Tasks and Actions it may be exposed to. By \textbf{Task}, we mean some non-trivial capability the agent must learn and may be quantitatively evaluated on. In Fig. 1, tennis and badminton are both Tasks. Tasks may also be subdivided into \textbf{Task Variants}; two Variants of the same Task retain similarity but are different enough to still pose difficulty in obtaining mastery of both. For example, “Tennis on a clay court at night” and “Tennis on a grass court during the day” might be two variants of the Tennis Task. For the purposes of this document, the delineation between tasks, and between variants of tasks is meant to be heuristic.

The \textbf{Lifetime} of an agent consists of the following stages:

\begin{enumerate}
    \item \textbf{Pre-deployment stage}: This is akin to genetic (or factory-installed) knowledge; the agent is endowed with pre-built knowledge for the given environment. The actual mechanism for the “endowment” may include meta-learning, genetic programming, pre-trained models based on statistically representative data, or other techniques for selecting a suitable “starting point” for the agent.
    
    \item \textbf{Deployment stage}: The agent is released into the environment and starts learning (see vertical dotted line in Fig. 1). The deployment stage consists of a sequence of Learning Experiences (LXs; more on this below) optionally interspersed with Evaluation Experiences (EXs). Each Learning or Evaluation Experience is drawn from an environment-specific Task. A Lifelong Learning Scenario (LLS), is a template for a particular sequence of learning experiences, intended to exercise one or more conditions of lifelong learning.
\end{enumerate}

Lifetimes are broken into smaller basic units we call Experiences. They come in two types: A \textbf{Learning Experience (LX)} is a minimum amount of experience with a Task that enables some learning activity on part of the agent; it is environment and task-specific. For example, in Arcade, a playthrough of a Breakout game would be a single Learning Experience. An \textbf{Evaluation Experience (EX)} is similar to a Learning Experience: it is an experience within a task used to evaluate an agent’s capabilities. Colloquially, the agent’s ability to learn is “turned off” during an EX.

Each Learning Experience will generate one or more environment and task-specific performance measures (e.g., in single playout of Arcade Breakout, these might be the total number of bricks destroyed and the total amount of time the ball is kept in motion). Over a lifetime, an agent may encounter tens of thousands to millions of LXs.

It is useful to estimate how much benefit a Lifelong Learning Agent has gained from being exposed to diverse Tasks during a Lifetime. To this end, we also consider \textbf{Single-Task Experts (STEs)}, which are L2 agents only ever exposed to a single Task after the pre-deployment training.

We assume that, in almost all cases of interest, an agent's learning will be stochastic, from some combination of its algorithmic core and its environmental exposures. Thus, a single lifetime will not not sufficient to characterize an agent’s lifelong learning capability. Reliably assessing the variability in the responses of the agent, assuming the inherent variability of its inputs (which could be expressed as statistical distributions,) requires assessing performance of the agent over multiple lifetimes.

Below is an illustrative approach:

\begin{enumerate}
    \item Exposure a blank state agent to a fixed Pre-Deployment process.
    \item Copy the agent's state after this Pre-Deployment, and independently exercise each copy on the same Lifelong Learning Scenario.
    \item Aggregate the metrics across the clones in a statistically meaningful manner.
\end{enumerate}

\subsection{Performance Measures and Metrics}\label{sec:measures}

An agent’s Learning and Evaluation Experiences are expected to generate a variety of domain, environment, and \textbf{task-specific performance measures}, only some of which may be relevant for lifelong learning. Data for performance measures are collected in the context of a learning scenario, a high-level description of what the agent will learn and the tasks it will experience. A Lifelong Learning Scenario is the instantiation of a learning scenario in a specific environment.

The Lifelong Learning metrics are scenario, domain, environment and task-agnostic numbers that characterize one or more \textbf{lifelong learning capabilities} across the lifetime of the agent. Each Learning Experience (LX) is also assumed to generate one or more scenario, domain, environment and application-specific performance metrics.

\begin{figure}
    \centering
    \includegraphics[width=0.65\textwidth]{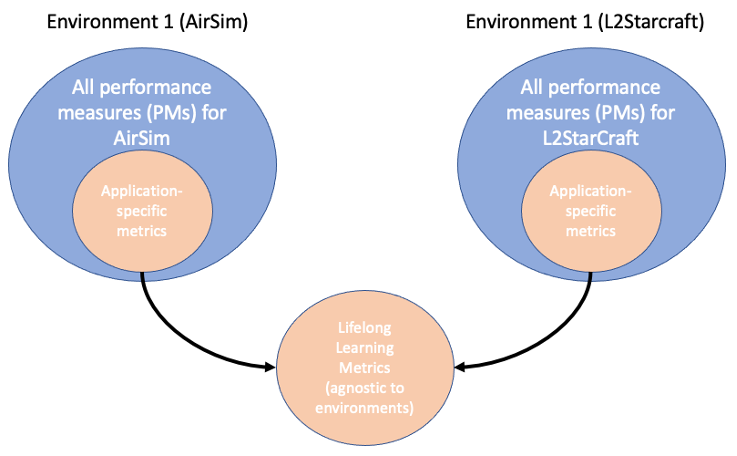}
    \caption{Environments such as AirSim and L2StarCraft define performance measures. Some subset of these are treated as Application-Specific Metrics (Section~\ref{sec:measures}), which feed into the calculation of Lifelong Learning Metrics (Section~\ref{sec:metrics}).}
    \label{fig:performance_measures}
\end{figure}

Fig.~\ref{fig:performance_measures} summarizes the relationship between performance measures, application-specific metrics, and lifelong-learning metrics. The following sections illustrate this analysis pipeline.

\subsection{Learning and Evaluation}

\begin{figure}
    \centering
    \includegraphics[width=0.75\textwidth]{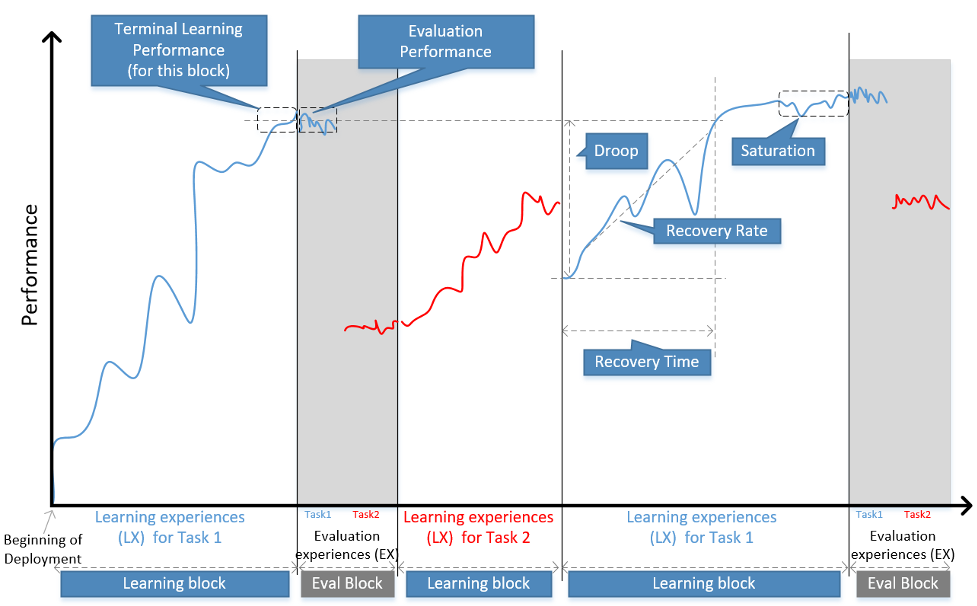}
    \caption{Illustration of terms related to metrics calculation in a single agent lifetime. A learning 
block may have thousands to millions of LXs. Note that Droop is relative to the most recent Terminal Learning Performance of the same task.
}
    \label{fig:sample_lifetime}
\end{figure}

Fig.~\ref{fig:sample_lifetime} illustrates some basic terms in setting up a Lifelong Learning Scenario and in the resulting metrics calculation. It assumes that the agent is deployed and has to learn two tasks, and that we are interested in how performance on the first task (Task-1) is affected by intervening experience with the second task (Task-2). The expected appearance and structure of these performance curves motivate the later development of our lifelong learning metrics.

One set of terms in Fig. 3 relates to how the learning scenario is sequenced. As noted earlier, the deployment stage consists of a sequence of Learning Experiences (LXs) optionally interspersed with \textbf{Evaluation Experiences (EXs)}. Each LX is for a Task, instantiated with some specific parameters.

Our assumption is that a single Learning Experience's granularity is too fine to enable reliable estimates of trends in agent learning capability. Thus, we adopt several less-granular categories. A set of contiguous LXs with the same Task and parameters is a \textbf{Learning Regime}. A sequence of Learning Experiences (potentially containing multiple Regimes) is a \textbf{Learning Block}, and a sequence of Evaluation Experiences is an \textbf{Evaluation Block}. Thus, a Lifelong Learning Scenario specifies a sequence of blocks.

There are three important points to keep in mind:

\begin{itemize}
    \item Performance measures need to be collected from Learning Blocks as well as Evaluation Blocks. This allows us to characterize how performance changed (i.e., recovered, dropped) within a Learning Block relative to prior Learning Blocks.
    \item Lifelong Learning Scenarios do not need to have Evaluation Blocks. For example, say we want to only assess the agent’s ability to learn continuously and adapt to changes in the environment. This might be done with a scenario that consists of a sequence of Learning Blocks (each Block boundary corresponds to a change in the environment), and we assess how well the agent recovers after each change.
    \item The notion of Learning Regimes and Blocks, Evaluation Blocks, and Tasks is agnostic to specific details of the agent and its environment. For example, the notion of a Learning Regime enables a task's definition to shift over the course of a lifetime, and our Lifelong Learning Metrics (Section~\ref{sec:metrics}) do not require that an agent be aware (or unaware) of the Task it is currently Experiencing. (A Task Identification score could be an Application-Specific Metric, in fact.) Thus, our framework includes existing Lifelong Learning scenarios like task-, domain-, and class-incremental learning~\cite{vandeVen2019ThreeScenarios}.
\end{itemize}

The other set of terms in Fig. 3 identify characteristics of the performance curves. The \textbf{Terminal Learning Performance (TLP)} is the performance at the end of a Learning Block; the \textbf{Evaluation Performance} is the performance on a task measured within an Evaluation Block, the \textbf{Droop} is the drop in performance relative to the most recent TLP for the same task; the \textbf{Recovery Time} is the number of LXs needed to get back to the most recent TLP (for the same task); and the \textbf{Recovery Rate} is a measure of how quickly that recovery happens. 

By default, the TLP is calculated by averaging performance over the last 10\% of the LXs in a Learning Block. When an Evaluation Block contains more than one EX (because of, e.g., stochasticity in the evaluation), the Evaluation Performance is the mean of all performance values in that EB.

\subsection{Application-Specific Metrics}

\subsubsection{Example of a single-task lifelong learning scenario}

\begin{figure}
    \centering
    \includegraphics[width=0.5\textwidth]{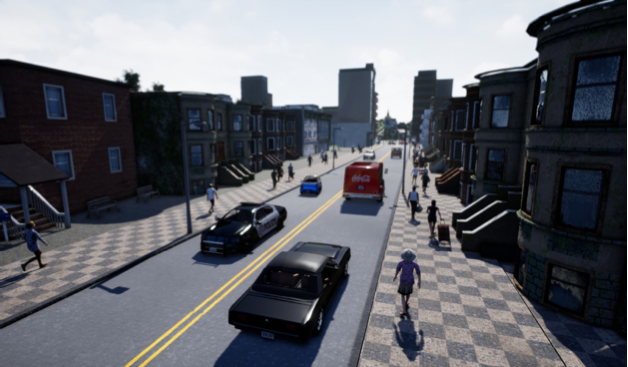}
    \caption{An example street view from the CARLA simulator. Image taken from~~\cite{Dosovitskiy17CARLA}.}
    \label{fig:carla}
\end{figure}

Table~\ref{table:single_task} illustrates an example Learning Scenario for assessing performance of agents within the CARLA (Fig.~\ref{fig:carla}) and SplitMNIST environments.

\begin{table}[htp!]
\begin{tabular*}{16.35cm}{|p{3cm}||p{6cm}|p{6cm}|}
 \hline
 & CARLA & SplitMNIST \\
 \hline\hline
 Task   
 & Task-1: Navigate from Point A to B in minimum time. 
 
 \begin{itemize}
     \item One LX = one end-to-end navigation. 
     \item Parametric variation = weather, time of day, etc.
     \item Regime 1: noonday lighting, foggy
     \item Regime 2: dusk/evening light, rainy
     \item Regime 3: stochastic time of day, sunny weather
\end{itemize}
 & 
 Task-1: Classify images as either \texttt{0} or \texttt{1} 
 
 \begin{itemize}
     \item One LX = minibatch of sixteen randomly sampled images
     \item Parametric variation = random brightness and contrast shifts
     \item Regime 1: All images brighter than normal
     \item Regime 2: All images darker than normal
     \item Regime 3: Images have the chance to have very high contrast
 \end{itemize}
\\\hline
 Lifelong Learning Scenario
 &
 \begin{itemize}
     \item Learning Block 1: 10000 LXs with Task-1 (regime 1)
     \item Learning Block 2: 5000 LXs with Task-1 (regime 2)
     \item Learning Block 3: 10000 LXs with Task-1 (regime 3)
     \item  (No Evaluation Blocks by design)
 \end{itemize}
 &
 \begin{itemize}
     \item Learning Block 1: 1000 LXs with Task-1 (regime 1)
     \item Learning Block 2: 500 LXs with Task-1 (regime 2)
     \item Learning Block 3: 1000 LXs with Task-1 (regime 3)
     \item (No Evaluation Blocks by design)
 \end{itemize}
 \\\hline

Application-specific metrics
&
Task-1. Weighted combination of time to destination and safe speed.
&
Task-1: Classification Accuracy\\\hline
Result of running Lifelong Learning Scenario
&
\begin{itemize}
    \item Learning Block 1: 10000 Task-1 reward values
    \item Learning Block 2: 5000 Task-1 reward values
    \item Learning Block 3: 10000 Task-1 reward values
\end{itemize}
&
\begin{itemize}
    \item Learning Block 1: 1000 Task-1 metrics
    \item Learning Block 2: 500 Task-1 metrics 
    \item Learning Block 3: 1000 Task-1 metrics
\end{itemize}
\\\hline\hline
\end{tabular*}
\caption{Sample collection of application-specific metrics for a single-task LLS}
\label{table:single_task}
\end{table}

There are three points to note in the above table:

\begin{enumerate}
    \item The above Lifelong Learning Scenarios start with the Deployment Stage
    \item 	The specific number of LXs may be different for each environment (e.g., 10000 LXs for Task-1 for CARLA; 1000 LXs for Task-1 for SplitMNIST). This is due to the intrinsic learning difficulty of each environment and the informativeness of each Learning Experience (i.e., how much it can contribute toward the agent’s learning).
    \item	Each LX generates an application-specific metric, and these metrics are environment and task-specific. For CARLA Task-1, the application-specific metric is a task-specific reward value, whereas for SplitMNIST Task-2, the application-specific metric is classification accuracy. These metrics may have very different magnitudes and ranges. (The task-specificity of the metrics is further illustrated in the next section, see Table 2).
\end{enumerate}

\begin{figure}[b!]
    \centering
    \includegraphics{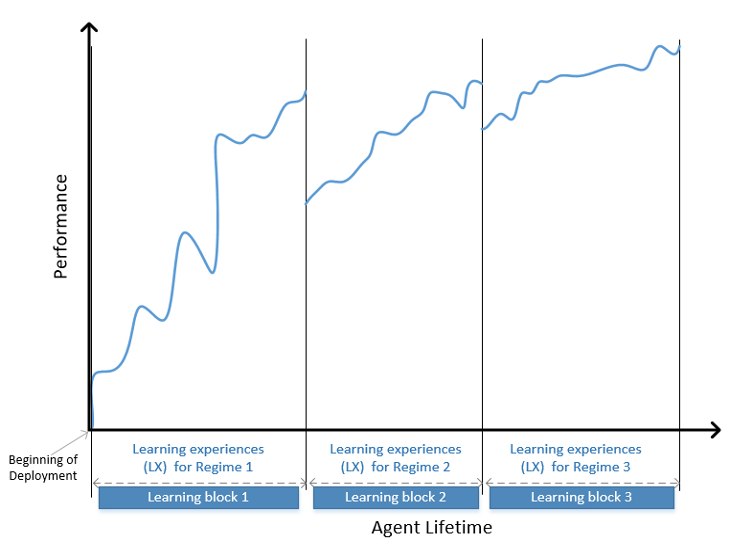}
    \caption{Notional performance time series for learning scenario in Table 1, for a single application-specific metric for a single environment.  All three Learning Blocks use the same Task but with different parameter values (regimes). Note that there are no Evaluation Blocks by design.}
    \label{fig:single_task}
\end{figure}

At the end of the above LLS for a learning agent, the result would be a time series (with 25000 points for CARLA, and 2500 points for SplitMNIST) that notionally looks like Fig.~\ref{fig:single_task}. We might then calculate the droop, recovery time, etc. for each Learning Block.

\subsubsection{Example of a multi-task lifelong learning scenario}

Table~\ref{tab:multi_task} illustrates an example Learning Scenario for assessing Transfer between two tasks by training agents within the CARLA and SplitMNIST environments. For a single LLS, the Learning and Evaluation Blocks are assumed to adhere to a fixed parameter regime.

\begin{table}[htp!]
    \centering
    \resizebox{0.85\textwidth}{!}{%
    \begin{tabular*}{16.35cm}{|p{3cm}||p{6cm}|p{6cm}|}
         \hline& CARLA & SplitMNIST\\\hline\hline
         Task
         &
         Task-1: Navigate from Point A to B in minimum time
         
         \begin{itemize}
             \item One LX = one end-to-end navigation process
             \item Parametric variation = weather, time of day, etc.
         \end{itemize}
         
         Task-2: Follow a designated vehicle
         
         \begin{itemize}
             \item One LX = following designated vehicle for a fixed amount of time.
             \item Parametric variation = weather, vehicle to be followed, distance to vehicle, etc
         \end{itemize}
         &
         Task-1: Classify images as either \texttt{0} or \texttt{1} 
         
         \begin{itemize}
             \item One LX = minibatch of sixteen randomly sampled images
             \item Parametric variation = random brightness and contrast shifts
         \end{itemize}
         
         Task-2: Classify images as either \texttt{2} or \texttt{3}
         
         \begin{itemize}
             \item One LX = minibatch of sixteen randomly sampled images
             \item Parametric variation = random brightness and contrast shifts
         \end{itemize}

        \\\hline
        Lifelong Learning Scenario
        &
        \begin{itemize}
            \item Evaluation Block 1: 100 EXs with Task-1, 100 with Task-2
            \item Learning Block 1: 5000 LXs with Task-1
            \item Evaluation Block 2: 100 EXs with Task-1, 100 with Task-2
            \item Learning Block 2: 10000 LXs with Task-2
            \item Evaluation Block 3: 100 EXs with Task-1, 100 with Task-2
        \end{itemize}
        &
        \begin{itemize}
            \item Evaluation Block 1: 10 EXs with Task-1, 10 with Task-2
            \item Learning Block 1: 1000 LXs with Task-1
            \item Evaluation Block 2: 10 EXs with Task-1, 10 with Task-2
            \item Learning Block 2: 2000 LXs with Task-2
            \item Evaluation Block 3: 10 EXs with Task-1, 10 with Task-2
        \end{itemize}
        \\\hline
        Application-specific metrics 
        &
        \begin{itemize}
            \item Task-1: Weighted combination of time to destination, safe speed, etc.
            \item Task-2: Weighted combination of distance to other vehicle, safe speed, etc.
        \end{itemize}
        &
        \begin{itemize}
            \item Task-1: Classification accuracy
            \item Task-2: Classification accuracy
        \end{itemize}
        \\\hline
        Result of running Lifelong Learning Scenario
        &
        \begin{itemize}
            \item Evaluation Block 1: 100 Task-1 \& 100 Task-2 reward values
            \item Learning Block 1: 5000 Task-1 reward values
            \item Evaluation Block 2: 100 Task-1 \& 100 Task-2 reward values
            \item Learning Block 2: 10000 Task-2 reward values
            \item Evaluation Block 3: 100 Task-1 \& 100 Task-2 reward values
        \end{itemize}
        &
        \begin{itemize}
            \item Evaluation Block 1: 10 Task-1 \& 10 Task-2 metrics
            \item Learning Block 1: 1000 Task-1 metrics (Application-specific metrics per LX) 
            \item Evaluation Block 2: 10 Task-1 \& 10 Task-2 metrics
            \item Learning Block 2: 2000 Task-2 metrics (Application-specific metrics per LX)
            \item Evaluation Block 3: 10 Task-1 \& 10 Task-2 metrics
        \end{itemize}
        \\\hline\hline
    \end{tabular*}
    }
    \caption{Sample collection of application-specific metrics for a multi-task LLS}
    \label{tab:multi_task}
\end{table}

Note that an LX may generate multiple application-specific metrics. Furthermore, the range of values for the application-specific metrics will vary across Tasks (across tasks within the same environment, and \emph{a fortiori}, across environments). For example, in StarCraft, “total resources collected” may range from 0 to 200; “\# of enemy units defeated” may range from 0 to 10. Hence, the values for application-specific metrics will not be directly comparable across Tasks without a normalizing function that can account for the difference in task difficulty.

\begin{figure}[b!]
    \centering
    \includegraphics{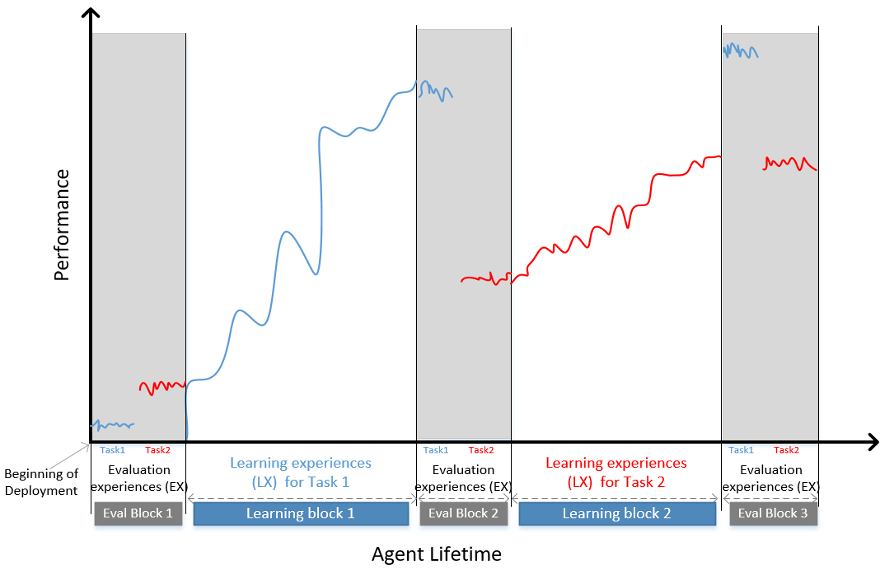}
    \caption{Notional performance time series for learning scenario in Table 2, for a single environment and a single application-specific metric per task within that environment.}
    \label{fig:multi_task}
\end{figure}

As output from the LLS for a learning agent, the result would be a time series that notionally looks like Fig.~\ref{fig:multi_task}. In this illustrative case, we might note that Task-2 performance increases between Eval Blocks 1 and 2 (indicating strong forward transfer), and Task-1 performance increases slightly between Eval Blocks 2 and 3 (indicating weak reverse transfer).

\subsubsection{Preprocessing application-specific metrics}

In order to be valid across variations in domain, scenario, and environment, the Lifelong Learning Metrics require Application-Specific Metrics to be preprocessed in certain ways. The primary requirement is that task performances be normalized to a fixed range. This accounts for the fact that different application-specific metrics might have naturally different ranges of variation. E.g., a classification task might have performance values based on a minibatch accuracy rate (ranging from 0 to 1), and a reinforcement learning task might have performance values based on a reward function that cluster in the range -100 to 1,000.

We have also found it useful to smooth performance curves (thereby mitigating some of the stochasticity of the learning process), and to clamp outlier performance values. In Appendix~\ref{sec:preprocessing}, we give a specific instantiation of this procedure, but the validity of the metrics does not rely on a particular preprocessing strategy.

Beyond any specific preprocessing scheme, we still recommend that metrics values be presented in the context of the scenarios in which they were generated. Improvements in values can be confounded by underlying difficulty of scenario. Such contextual descriptions are essential to prevent gaming or unfair comparisons of the metrics.

\subsection{Generating Lifelong Learning Metrics}

The lifelong learning metrics are domain, scenario, environment and task-agnostic numbers that characterize the one or more \textbf{lifelong learning capabilities across the lifetime of the agent}. In other words, \textbf{each Lifelong Learning metric has a single value per lifetime of an agent}. Some lifelong learning metrics will first produce a set of task-specific values; task-specific metrics are aggregated into a single value via the mean operation at the end of the agent’s lifetime.

Notionally, each agent lifetime generates a sequence of application-specific metrics; these metrics feed into the computation of the lifelong learning metric, which generates a single environment-agnostic value.

\section{Metrics for Lifelong Learning}\label{sec:metrics}

\subsection{Proposed Lifelong Learning Metrics}

Table~\ref{tab:metrics} presents each metric in a jargon-free manner. The L2 Conditions are defined in Section~\ref{sec:conditions}. Several additional metrics are presented in Appendix~\ref{sec:supplemental}.

\begin{table}[h!]
    \centering
    \resizebox{0.95\textwidth}{!}{%
    \begin{tabular*}{16.35cm}{|p{6cm}||p{4cm}|p{5cm}|}
         \hline
         \textbf{Metrics Assesses System's Ability To:}
         & 
         \textbf{L2 Condition}
         &
         \textbf{Metric Name:}
         \\\hline\hline
         Be robust to catastrophic forgetting despite the introduction of new parameters and/or tasks
         &
         Continual Learning
         & 
         Performance Maintenance
         \\\hline
         Make use of knowledge acquired from one task to catalyze learning a new task
         &
         Transfer and Adapation
         & 
         Forward Transfer
         \\\hline
          Make use of knowledge acquired from a new task to improve performance on a previously learned task
         &
         Transfer and Adaptation
         & 
         Backward Transfer
         \\\hline
         Approach or exceed the performance of a Single Task Expert (STE)
         &
         Transfer and Adaptation
         &
         Performance Relative to STE
         \\\hline
         Make use of incoming knowledge to learn tasks quickly and efficiently
         &
         Sustainability
         &
         Sample Efficiency\\\hline
    \end{tabular*}
    }
    \label{tab:metrics}
    \caption{Lifelong Learning Metrics}
\end{table}

\subsection{Single-Task Metrics, Cross-Task Metrics, and Aggregation Across a Lifetime}

Of the proposed Lifelong Learning Metrics, some are calculated using performance measures for a single Task, and others are calculated between pairs of Tasks. Furthermore, most metrics utilize quantities calculated multiple times during a lifetime: Performance Maintenance, for example, uses Maintenance Values calculated after each Learning Block.

To report on a system’s lifetime performance, each metric calls for its values to be aggregated into a single quantity. We recommend performing this aggregation with the mean operator, although other operators are possible.\footnote{Due to computational constraints, lifelong learning scenarios may feature fairly short lifetimes (e.g., three repetitions of four core tasks). In this low-data setting, distribution means are often easier to estimate than distribution medians.}  It can also be informative to examine how these sub-metrics change over time; this can inform understanding of how an agent’s learning dynamics vary.

\subsection{Contrasts vs. Ratios for Calculating Transfer}

Our formulation of forward and backward transfer uses the contrast function:

$$\mathrm{Contrast}(a, b) = \frac{a-b}{a+b}$$

The contrast function is qualitatively similar to the ratio function: $\mathrm{Ratio}(a,b)= \dfrac{a}{b}$. However, contrasts are defined when $b = 0$, unlike ratios. This makes them more robust in our use case, when application-specific metrics could be zero. Interpreting a ratio is easier than interpreting a contrast, so ratios may also be calculated if desired.

\subsection{Proposed Lifelong Learning Metrics}

An important note about these metric scores is that they must be presented in the context of the scenarios in which they were generated.  For example, an agent might exhibit a reduced positive PR score if the input conditions are such that the agent has to “work harder” to achieve improved performance over the Learning Experiences. Such contextual descriptions are essential to prevent gaming or unfair comparisons of the metrics.

For each metric, we describe how to qualitatively interpret the values of the metric: does the metric suggest a system is demonstrating lifelong learning? To ensure the meaningfulness of these interpretations, they should be considered in the context of the learning scenario that generated the metrics. A poorly-designed scenario can yield nonsensical metric values. 

For each metric, we explain how to calculate it, and give an interpretation of its values (i.e., does a given observed value indicate that a system is demonstrating lifelong learning). Then we also note references from the literature that use this metric or something similar to it. For example, transfer has been formalized in several different ways. These references were sourced from papers published at NeurIPS 2020 and 2019, ICML 2020 and 2019, and ICLR 2020 and 2019.

\begin{figure}
    \centering
    \includegraphics[width=0.95\linewidth]{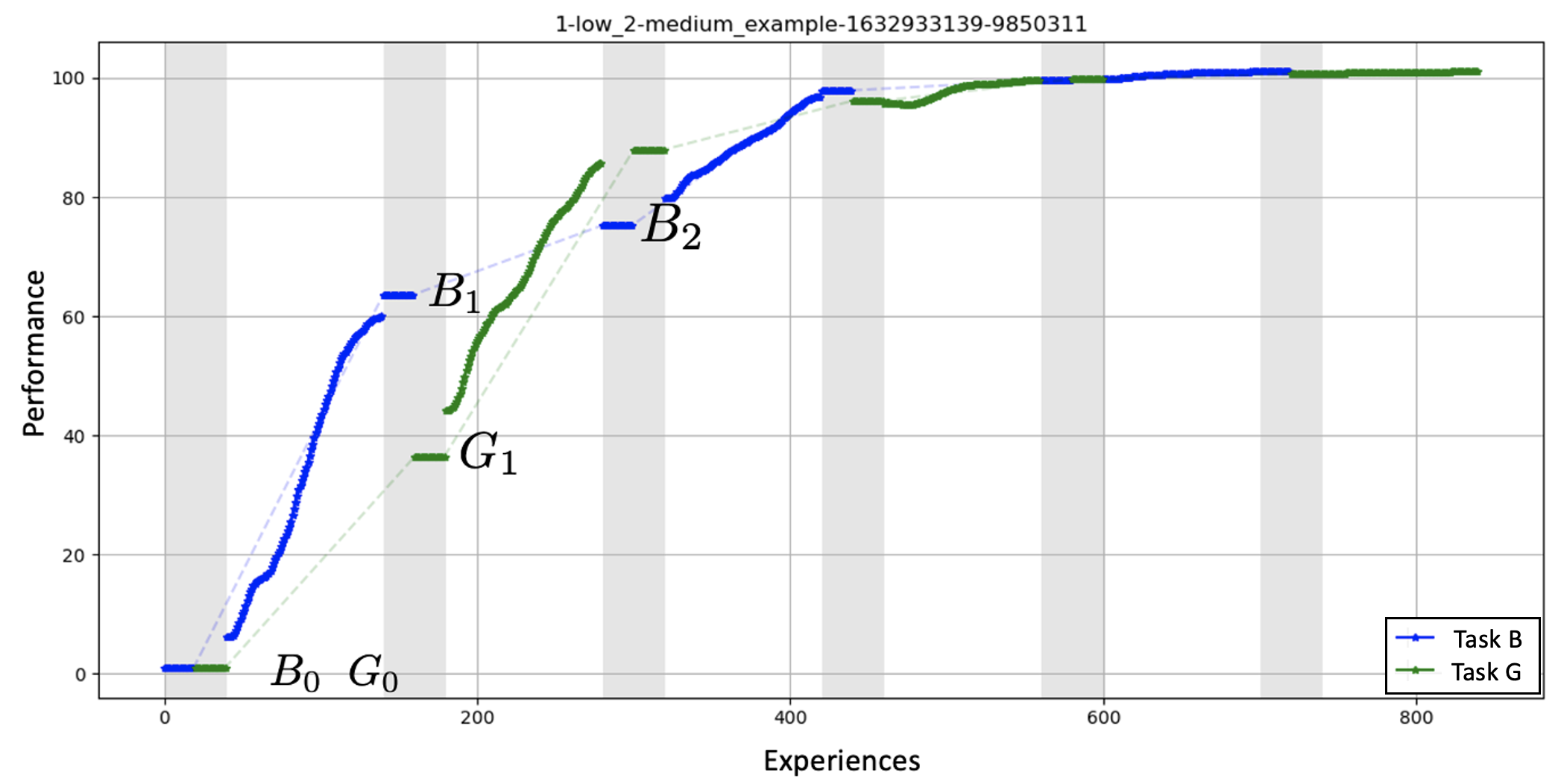}
    \caption{Performance curves for an L2 agent in a scenario with two tasks. White regions indicate Learning Blocks, and shaded regions indicate Evaluation Blocks. $B_i$ and $G_i$ refer to performance in the $i$th Evaluation Block for the blue and green tasks, respectively.}
    \label{fig:perf_A}
\end{figure}

\begin{figure}
    \centering
    \includegraphics[width=0.95\linewidth]{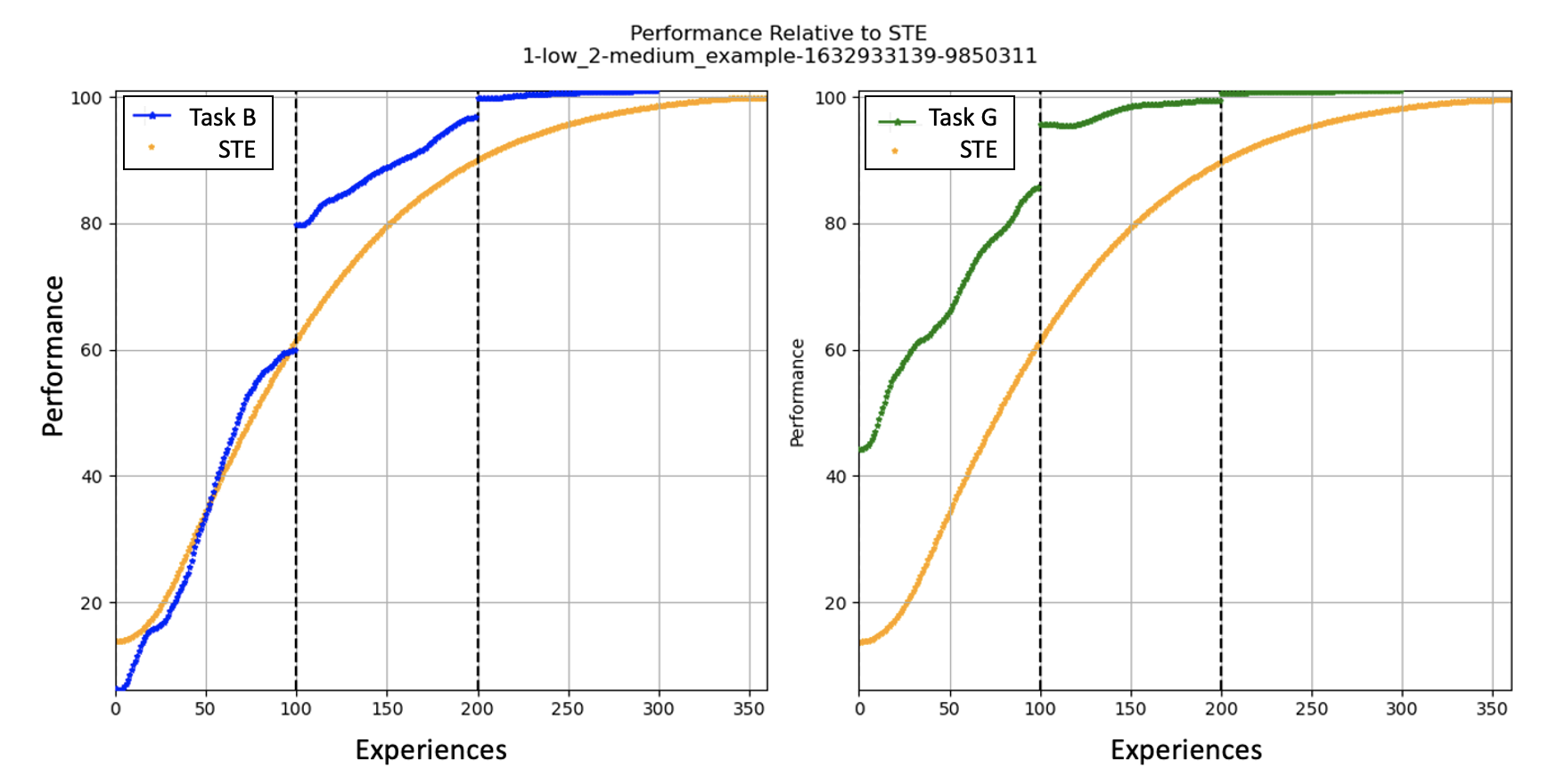}
    \caption{Single-task expert (orange) and L2 agent (blue and green) curves for the scenario shown in Fig.~\ref{fig:perf_A}. Dashed lines indicate the boundaries between Learning Blocks.}
    \label{fig:ste_A}
\end{figure}

\begin{figure}
    \centering
    \includegraphics[width=0.95\linewidth]{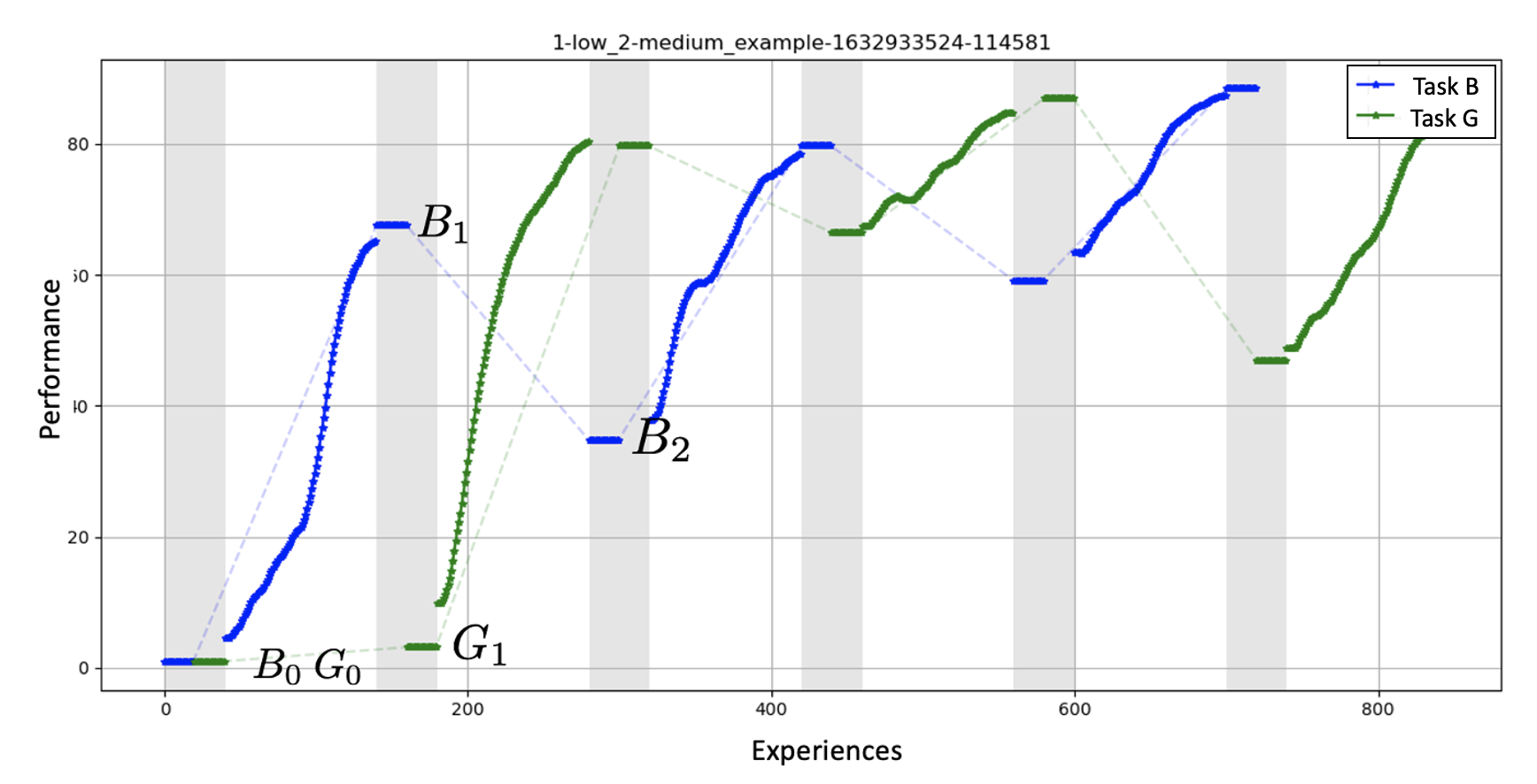}
    \caption{Performance curves for an L2 agent in a scenario with two tasks. White regions indicate Learning Blocks, and shaded regions indicate Evaluation Blocks. $B_i$ and $G_i$ refer to performance in the $i$th Evaluation Block for the blue and green tasks, respectively.}
    \label{fig:perf_B}
\end{figure}

\begin{figure}
    \centering
    \includegraphics[width=0.95\linewidth]{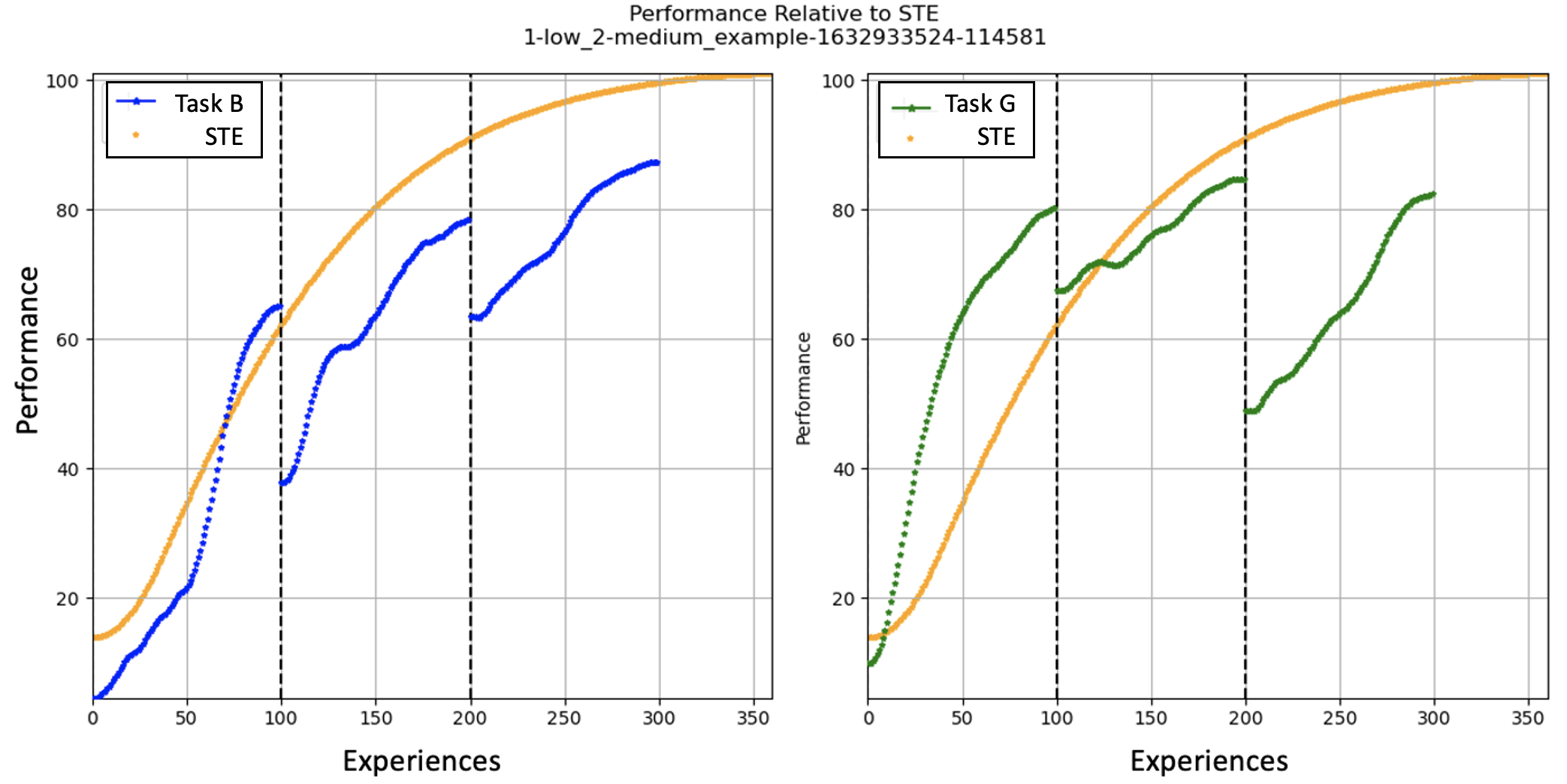}
    \caption{Single-task expert (orange) and L2 agent (blue and green) curves for the scenario shown in Fig.~\ref{fig:perf_B}. Dashed lines indicate the boundaries between Learning Blocks.}
    \label{fig:ste_B}
\end{figure}

In Figs.~\ref{fig:perf_A},~\ref{fig:ste_A},~\ref{fig:perf_B}, and~\ref{fig:ste_B}, we show performance curves for two representative scenarios, generated by an idealized learning system. Figs.~\ref{fig:perf_A} and~\ref{fig:perf_B} show the performance of the L2 agent in Learning and Evaluation Blocks, and Figs.~\ref{fig:ste_A} and~\ref{fig:ste_B} compare the Learning Block performance of the L2 agents to STEs.


\subsubsection{Performance Maintenance (PM)}

Does an L2 agent catastrophically forget a previously learned task? Metrics similar to PM are used in~\cite{Mendez2020,Jung2020Adaptive,Joseph2020metaconsolidation}

\begin{itemize}
    \item Computed by: 

    \begin{itemize}
        \item Select an application-specific metric to monitor for the given environment (e.g., total reward).
        \item Set up a learning scenario with a sequence of Learning Blocks alternating with Evaluation Blocks. Each Evaluation Block exercises all the previously learned tasks.
        \item For a given Task:
        \begin{itemize}
            \item Calculate the Maintenance Values, defined as the differences between each of the Task's Evaluation Performances (excluding Evaluation Blocks immediately following the Task's Learning Blocks) and the Evaluation Performance after the Task's most recent Learning Block.
        \end{itemize}
        \item Performance Maintenance (PM) for a lifetime = mean Maintenance Value across the lifetime.
    \end{itemize}

    \item Interpretation of the metric:

    \begin{itemize}
        \item PM = 0: No forgetting; no additional learning.
        \item PM $>$ 0: (Demonstrates LL) that performance on task is getting better over lifetime; May be an indication of transfer.
        \item PM $<$ 0: (Does not demonstrate LL) Indicates forgetting.
    \end{itemize}

    \item Notes:

    \begin{itemize}
        \item For tasks with significant overlap, Performance Maintenance may act as an additional measure of transfer; however, even learning disjoint tasks should result in a minimum score of zero for PM.
        \item Performance Maintenance is confounded in lengthy scenarios where the system quickly saturates in task performance. As the scenario length increases, PM will trend toward 0, as no further improvement is possible.
        \item Fig.~\ref{fig:perf_A} shows positive Performance Maintenance for both the blue and green tasks.
        \item Fig.~\ref{fig:perf_B} shows negative Performance Maintenance for both the blue and green tasks.
    \end{itemize}
\end{itemize}

\subsubsection{Forward Transfer (FT)}

Does an L2 agent improve learning on a new task by leveraging data from previous tasks? Existing literature often calls this a “jumpstart” formulation of the concept of forward transfer, which measures the boost in performance one task gives to another when the second task is first seen. Metrics similar to FT are used in~\cite{Mendez2020, Pan2021continual,Ke2020MixedSequence,Rolnick2019Experience}

\begin{itemize}
    \item Computed by: 

    \begin{itemize}
        \item Select an application-specific metric to monitor for the given environment (e.g. total reward).
        \item Set up a learning scenario beginning with initial Evaluation Blocks for all tasks, followed by a sequence of Learning Blocks (for different tasks) alternating with Evaluation Blocks.
        \item Assuming a block sequence like: Eval Block 1, Learning Block 1 (Task-1), Eval Block 2, then Task-2’s Forward Transfer (from Task-1) is computed as the contrast of the Evaluation Performances of Task-2 in Eval Block 2 to Eval Block 1.
        \item The Forward Transfer for a lifetime is the mean of each task pair’s first Forward Transfers.
    \end{itemize}

    \item Interpretation of metric:

    \begin{itemize}
        \item FT = 0: No transfer or forgetting.
        \item FT $>$ 0: (Demonstrates LL) Indicates positive forward transfer.
        \item FT $<$ 0: (Does not demonstrate LL) Indicates interference.
    \end{itemize}

    \item Notes:

    \begin{itemize}
        \item FT is computed only once per task pair, and can only be computed before that task has been learned, since a task is only “new” before it has been seen.
        \item Order of the tasks is important. Forward Transfer may be present from Task 1 $\to$ 2, but not 2 $\to$ 1 or vice versa.
        \item The usefulness of a task’s FT score depends on the structure of the learning scenario. If, in Fig. 9’s example below, Task-1’s first learning block were very short in length, then the calculated transfer from Task-1 to Task-2 would not be an informative metric, as there would not been sufficient experience for an agent to learn a shared task representation.
        \item As a ``jumpstart'' metric, FT is highly dependent on a task's initial performance.
        \item In Figs.~\ref{fig:perf_A} and~\ref{fig:perf_B}, the forward transfer from the blue task (\texttt{Task B}) to the green task (\texttt{Task G}) is given by $\text{Contrast}(G_1, G_0) = \dfrac{G_1 - G_0}{G_1 + G_0}$.
    \end{itemize}
\end{itemize}

\subsubsection{Backward Transfer (BT)}

Does an L2 agent improve performance on a previously learned task by leveraging data from new tasks?  This is a “jumpstart” formulation of the concept of backward transfer, which measures the boost in performance one task gives to another when the second task is first seen. Metrics similar to BT are used in~\cite{Pan2021continual,Gupta2020,Ke2020MixedSequence,Ebrahimi2020Uncertainty-guided}

\begin{itemize}
    \item Computed by:

    \begin{itemize}
        \item Select an application-specific metric to monitor for the given environment (e.g., total reward).
        \item Set up a learning scenario with a sequence of Learning Blocks. Between each Learning Block, there are Evaluation Blocks for each of the other tasks.
        \item For each task:
        \begin{itemize}
            \item Compute the Backward Transfer Contrast, defined as the contrast of the average performance within the most recent Evaluation Block to the second-most recent Evaluation Block.
            \item 	Backward Transfer for task T = the average of the Backward Transfer Contrasts.
        \end{itemize}
        \item The Backward Transfer for a scenario is the mean of each task pair’s first calculated Backward Transfer value.
    \end{itemize}

    \item Interpretation of metric:
    
    \begin{itemize}
        \item 	BT = 0: No interference, but no transfer, either.
        \item 	BT $>$ 0: (Demonstrates LL) Indicates positive backward transfer.
        \item BT $<$ 0: (Does not demonstrate LL) Indicates interference.
    \end{itemize}

    \item Notes:
    
    \begin{itemize}
        \item The scenario’s Backward Transfer value takes as input only each task’s first calculated BT value. However, unlike Forward Transfer, Backward Transfer can be computed several times for each task in a lifetime. 
        \item These additional BT values may still be calculated and track to provide additional mechanisms for assessing performance. Note that, if a lifetime lasts sufficiently long that agents saturate in all their tasks, BT values may trend toward 0.
        \item While there is no need to start the scenario with a “pop quiz” evaluation, both LX and EX blocks of the same task are required to compute BT.
        \item The usefulness of a task’s BT score depends on the structure of the learning scenario. If, in Fig. ~\ref{fig:perf_A}’s example above, Task-2’s learning block were very short in length, then the calculated transfer from Task-2 to Task-1 would not be an informative metric, as there would not been sufficient experience for an agent to learn a shared task representation.
        \item As a ``jumpstart'' metric, BT is highly dependent on a task's initial performance.
        \item In Figs.~\ref{fig:perf_A} and ~\ref{fig:perf_B}, the backward transfer from the green task (Task G) to the blue task (\texttt{Task B}) is given by $\text{Contrast}(B_2, B_1) = \dfrac{B_2 - B_1}{B_2 + B_1}$.
    \end{itemize}
\end{itemize}

\subsubsection{Performance Relative to a Single-Task Expert (RP)}

How does the performance of an L2 system compare to a non-lifelong learner with comparable learning experience? Metrics similar to RP are used in~\cite{Joseph2020metaconsolidation}.

\begin{itemize}
    \item Computed by: 

    \begin{itemize}
        \item Select an application-specific metric to monitor for the given environment (e.g. total reward) that has also been logged for a Single Task Expert (e.g. a non-lifelong learner).
        \item Set up a learning scenario with some sequence of Learning Blocks for some number of tasks.
        \item For a given task $T$:
        \begin{itemize}
            \item Extract the Learning Blocks for Task $T$, in order of appearance. This is illustrated in Figs.~\ref{fig:ste_A} and~\ref{fig:ste_B}.
            \item Compare the “area under the curve” for the lifelong learner experiencing Task $T$ with the area under the curve for the equivalent Single Task Expert.
            \item Formally, Compute the Single Task Expert Ratio, defined as the ratio of the sum of the application-specific metric over all of the Learning Experiences in the lifetime to the sum of the same application-specific metric over the same amount of learning experiences for the Single Task Expert.
            \item Relative Performance for Task $T$ = the Single Task Expert Ratio.
        \end{itemize}
        \item The RP for a lifetime is the mean of each task’s RP score.
    \end{itemize}

    \item Interpretation of metric:

    \begin{itemize}
        \item RP = 1 : L2 agent demonstrates the same overall performance compared to the Single Task Expert’s performance.
        \item  	RP $>$ 1: (Demonstrates LL) L2 agent demonstrates better performance given the same amount of Learning Experiences than the Single Task Expert.
        \item RP $<$ 1: (Does not demonstrate LL) L2 agent demonstrates worse performance given the same amount of Learning Experiences than the Single Task Expert.
    \end{itemize}
    
    \item Notes:

    \begin{itemize}
        \item For each task, Single Task Expert “learning curves” are required.
        \item If the STE and L2 Learner learning curves contain differing numbers of LXs, RP is calculated by using the minimal number of LXs.
        \item 	No Evaluation Blocks are necessary for this computation.
        \item In Fig.~\ref{fig:ste_A}, RP is positive, and, in Fig.~\ref{fig:ste_B}, RP is generally negative.
    \end{itemize}
\end{itemize}

\subsubsection{Sample Efficiency (SE)}

Does the system learn quickly and efficiently over the course of its lifetime? Metrics similar to SE are used in~\cite{Rolnick2019Experience}.

\begin{itemize}
    \item Computed by: 

    \begin{itemize}
        \item Select an application-specific metric to monitor for the given environment (e.g. total reward) that has also been logged for a Single Task Expert (e.g. a non-lifelong learner).
        \item Set up a learning scenario with some sequence of Learning Blocks for some number of tasks.
        \item Intuitively, compare the saturation value of the Single Task Expert with that of the lifelong learner. 
        \item For each task $T$:
        \begin{itemize}
            \item Extract the Learning Blocks for Task $T$, in order of appearance. This is illustrated in Figs.~\ref{fig:ste_A} and~\ref{fig:ste_B}.
            \item Compute Saturation Value (the max of the rolling average of the application-specific metric) and the Experience to Saturation (the number of Learning Experiences it takes to achieve the Saturation Value) for both the L2 agent and the Single Task Expert (STE) system.
            \item Compute the ratio of the Saturation Values of the L2 agent and the STE.
            \item 	Compute the ratio of the Experience to Saturation (ETS) for the STE and the L2 Agent.
            \item Sample Efficiency = Saturation Value Ratio $\times$ Experience to Saturation Ratio.
            \item In Fig.~\ref{fig:ste_A}, SE is above 1, and, in Fig.~\ref{fig:ste_B}, SE is below 1.
        \end{itemize}
        \item The Sample Efficiency for a lifetime is the mean of each task’s SE score.
    \end{itemize}

    \item Interpretation of metric:

    \begin{itemize}
        \item SE = 1 : L2 agent demonstrates the same overall performance compared to the Single Task Expert’s performance.
        \item SE $>$ 1: (Demonstrates LL) L2 agent demonstrates better performance than the Single Task Expert; either it learns faster or achieves a higher overall value compared to the Single Task Expert.
        \item SE $<$ 1: (Does not demonstrate LL) L2 agent demonstrates worse performance than the Single Task Expert; either it learns slower or achieves a lower overall value compared to the Single Task Expert.
    \end{itemize}

    \item Notes:

    \begin{itemize}
        \item Single Task Expert “learning curves” are required for this computation
        \item Only tasks which demonstrate a steady-state/saturated performance are eligible for this metric.
    \end{itemize}
\end{itemize}

\section*{Acknowledgements}

Primary development of this document was funded by the DARPA Lifelong Learning Machines (L2M) Program. The authors would like to thank the performers in this program, particularly the members of the Definitions and Metrics Working Groups, who helped to develop and refine these metrics. Thanks also goes to the DARPA SETAs for this program -- Conrad Bell, Robert McFarland, Rebecca McFarland, and Ben Epstein.

\section*{Disclaimer}

The views, opinions, and/or findings expressed are those of the author(s) and should not be interpreted as representing the official views or policies of the Department of Defense or the U.S. Government.

\bibliographystyle{plain}
\bibliography{references}

\appendix

\section{Preprocessing Application-Specific Metrics}\label{sec:preprocessing}

Here, we give an example metrics preprocessing strategy. Note both the ordering of operations, and the specific values used for clamping and normalization. This strategy is intended for the case that there are multiple available performance curves for both L2 and STE agents, but it still works when there is only one curve of each agent type.

\begin{enumerate}
    \item For each Learning Block (across all available performance curves), smooth the performance curves with a flat window of length $L$, where $L$ is the minimum of 20\% of the LB's length (in LXs) and 100 LXs. Pad the smoothed curve so that it is the same length as the raw curve.
    \item For each Task Variant in the scenario, calculate the 10\% and above the 90\% percentiles of the distribution of that Task Variant's Application-Specific Metric Values, over the Lifelong Learning Agent and the STE.
    \item Reduce the impact of outliers by clamping values below the 10\% and 90\% percentiles to those values, and then scale all performance values to a fixed range. (We recommend 1 to 101, as having a minimum performance of 0 can cause jumpstart metrics to take uninformatively large values.)
    \item If a Task Variant has multiple known Single Task Experts, aggregate relevant Lifelong Learning Metrics by computing those metrics with respect to each Single Task Expert, then averaging the values at the end.
\end{enumerate}


\section{Supplemental Metrics}\label{sec:supplemental}

The following metrics have also been proposed, but they are not as fine-tuned in definition.

\subsection{Performance Recovery (PR)}

How effectively can an L2 agent’s performance ``bounce back'' after a change is introduced to its environment? Performance Recovery is a Single-Task metric.

\begin{itemize}
    \item Computed by:

    \begin{itemize}
        \item For each Task:
        \begin{itemize}
            \item Select an application-specific metric to monitor for the given environment (e.g. total reward).
            \item Set up a learning scenario with a sequence of LX blocks. Each LX block introduces a parametric change to an already-learned task.
            \item From the second Learning Block onwards, calculate the Recovery Time relative to the most recent Terminal Learning Performance. The “Recovery Time” is the number of LXs for performance to return to the previous Terminal Learning Performance.
            \item 	Task-Specific Performance Recovery = negative slope of the line of (Learning Block index, recovery time) values.
            \item Report Lifetime PR as the mean of all Task-Specific PRs.
        \end{itemize}
    \end{itemize}
    
    \item Interpretation:
    
    \begin{itemize}
        \item PR = 0: system is not exhibiting learning (the Recovery Times are not getting better or worse over the lifetime of the agent).
        \item PR $>$ 0: (Demonstrates LL) Recovery Times are decreasing over the lifetime, i.e., the system gets faster at recovering from environmental changes.
        \item PR $<$ 0: (Does not demonstrate LL) Recovery Times are increasing over the lifetime, i.e., the system gets slower at recovering from environmental changes. 
    \end{itemize}
    
    \item Notes:
    
    \begin{itemize}
        \item Performance Recovery can only be assessed while the system is learning, as a frozen system will not recover.
        \item If, during a Learning Block, the system never recovers to its most recent Terminal Learning Performance, the recovery time is (\# LXs in the learning block) + 1.
        \item 	If, due to transfer, a system’s performance starts above the previous Terminal Learning Performance, the recovery time is 0.
        \item To handle potentially infinite recovery times, we recommend the Theil-Sen estimator of slope~\cite{Sen1968}.
        \item For long lifetimes in which agent performance saturates, PR may trend to 0, as all of a system’s recovery times trend to 0.
        \item The PR slope is negative so that all L2M metrics have the same monotonic interpretation (i.e., bigger is better).
        \item This metric has been applied in evaluations, unlike the other supplemental metrics. However, experience in these suggested that it has high variability and limited utility and informativeness.
    \end{itemize}
\end{itemize}

\subsection{Cumulative Gain of a Lifelong Learner (CG)}

Does an L2 system continue to learn as new tasks are introduced over the course of its lifetime?

\begin{itemize}
    \item Computed by: 

    \begin{itemize}
        \item 	Select an application-specific metric to monitor for the given environment (e.g. total reward).
        \item Set up a learning scenario with some sequence of LX blocks for some number of tasks.
        \item For each LX block, compute a trend line through the application-specific metric to determine gain:
        \begin{itemize}
            \item If trendline has positive slope, gain = 1.
            \item 	If trendline has zero slope, gain = 0.
            \item 	If trendline has negative slope, gain = -1.
        \end{itemize}
        \item CG = average of all gains.
    \end{itemize}
    
    \item Interpretation of metric:
    
    \begin{itemize}
        \item CG $\geq$ 0: (Demonstrates LL)  L2 agent is not losing the ability to learn.
        \item CG $<$ 0: (Does not demonstrate LL) L2 agent’s ability to learn is deteriorating over time.
    \end{itemize}
    
    \item Notes:
    
    \begin{itemize}
        \item 	No EX blocks are required for this metric
    \end{itemize}
\end{itemize}

\subsection{Learn Burn (LB)}

Does an L2 system efficiently make use of learning opportunities when changes are introduced?

\begin{itemize}
    \item Computed by: 
    \begin{itemize}
        \item Select an application-specific metric to monitor for the given environment (e.g. total reward).
        \item Set up a learning scenario with some sequence of LX blocks for some number of tasks and/or parametric variations.
        \item 	After each change is introduced:
        \begin{itemize}
            \item Compute Burn Rate for change i, ($BR_i$), defined as the slope of the learning curve in an LX block following the change.
        \end{itemize}
        \item Compute Average Learn Rate across the whole lifetime, defined as the slope of the trendline across all performances.
        \item Learn Burn LB = average Burn Rate / Average Learn Rate.
    \end{itemize}

    \item Interpretation of metric:

    \begin{itemize}
        \item LB $>$ 1: (Does not demonstrate LL) L2 agent, on average, learns slowly while adapting to a change.
        \item LB $<$ 1: (Demonstrates LL)  L2 agent, on average, learns quickly while adapting to a change.
    \end{itemize}

    \item Notes:

    \begin{itemize}
        \item 	No EX blocks are required for this metric.
    \end{itemize}
\end{itemize}

\section{Glossary of Terms}\label{sec:glossary}

Table~\ref{tab:glossary} defines the terms used in this document.

\begin{table}[h!]
    \centering
    \begin{tabular*}{16.8cm}{|p{5cm}|p{11cm}|}
         \hline 
         Lifelong Learning Machine (equivalently, L2 System or L2 Agent)
         &
        The system that is capable of demonstrating lifelong learning in a specific domain and environment.
        \\\hline
        Condition of Lifelong Learning
        &
        A necessary characteristic of lifelong learning; something that a L2 agent exhibits. The set of L2 conditions comprises a necessary and sufficient definition of L2.
        \\\hline
        Domain
        &
        The kind of challenge being addressed by the lifelong learning agent.

        Examples: autonomous navigation; robotics; embodied agent
        \\\hline
        Environment
        &
        The computational setting for the domain; the “world” in which the L2 agent acts.  A given environment can support different domains (depending on how Tasks are formulated). Examples: StarCraft~\cite{StarCraft}, AirSim~\cite{Airsim2017fsr}, CARLA~\cite{Dosovitskiy17CARLA}, Habitat~\cite{habitat19iccv}, Arcade~\cite{bellemare13arcade}
        \\\hline
        \multicolumn{2}{|c|}{\textbf{Tasks and Learning Scenarios}}
        \\\hline
        Task
        &
        A single environmentally-relevant capability that a L2 Agent must learn; equivalently, a capability that is organized to achieve a single specific goal. 
        
        A Task is uniquely specified by an environment; observation space; action space; a reward function; and a range of parametric variations (that can be either goal-relevant or irrelevant).
        
        A Task may have a well-defined end state (e.g., agent “dies”; goal accomplished) or it may continue indefinitely (e.g, open-world exploration).
        
        Example: In Arcade, Pong is a Task with a potential range of variation (paddle width, ball speed, background color, etc.).
        \\\hline
        Task Instance
        &
        A instance of a Task; a specific environmental context that the L2 agent must act in to accomplish a goal. 
        
        Example: In Arcade, A Task Instance of Pong is an actual game that a L2 agent must play (with specific instantiated values of paddle width, ball speed etc.)
        \\\hline
        Learning Experience
        &
        An interaction with the environment that gives the L2 agent an opportunity to do task-relevant learning. 
        \\\hline
        Learning Scenario
        &
        A template for a sequence of Learning Experiences that exercises a L2 agent and allows it to exhibit one or more L2 conditions. A Learning Scenario is agnostic to specific Environments and Tasks.
        
        A Learning Scenario consists of an informal description of:
        
        1. The Task sequence encountered by the agent (e.g., how long the L2 agent experiences each Task; when new Tasks or new Task Instances are introduced).
        
        2. When and how the L2 agent is evaluated during that sequence
        
        3. What constitutes successful lifelong learning within that scenario
        
        A Learning Scenario supports one or more Metrics (i.e., a Learning Scenario allows for calculation of the metrics, and thereby assess the L2 capability of the agent).
        \\\hline
        Lifelong Learning Scenario
        &
        An instantiation of a Learning Scenario in a specific environment. Consists of a sequence of Task Instances drawn from one or more Tasks.
        \\\hline
        \multicolumn{2}{|c|}{\textbf{L2 Agent "Lifetime"}}
        \\\hline
        Pre-Deployment Stage
        &
        An optional stage where agents are endowed with pre-built knowledge or capabilities for a specific environment. Learning during this phase is not considered part of lifelong learning as such.
        \\\hline
        Deployment Stage
        &
        Once the agent is released into the field, this is considered the start of the agent’s “lifetime”, and the start of its lifelong learning. Consists of Learning and Evaluation Experiences.
        \\\hline
        Agent Lifetime
        &
        A sequence of one or more Lifelong Learning Scenarios.
        \\\hline
    \end{tabular*}
    \label{tab:glossary}
    \caption{Glossary of Terms}
\end{table}

\end{document}